# A Data-driven Dynamic Temporal Correlation Modeling Framework for Renewable Energy Scenario Generation

Xiaochong Dong, *Member, IEEE*, Yilin Liu, *Student Member, IEEE*, Xuemin Zhang, *Member, IEEE*, Shengwei Mei, *Fellow, IEEE*

*Abstract* — Renewable energy power is influenced by the atmospheric system, which exhibits nonlinear and time-varying features. To address this, a dynamic temporal correlation modeling framework is proposed for renewable energy scenario generation. A novel decoupled mapping path is employed for joint probability distribution modeling, formulating regression tasks for both marginal distributions and the correlation structure using proper scoring rules to ensure the rationality of the modeling process. The scenario generation process is divided into two stages. Firstly, the dynamic correlation network models temporal correlations based on a dynamic covariance matrix, capturing the time-varying features of renewable energy while enhancing the interpretability of the black-box model. Secondly, the implicit quantile network models the marginal quantile function in a nonparametric, continuous manner, enabling scenario generation through marginal inverse sampling. Experimental results demonstrate that the proposed dynamic correlation quantile network outperforms state-of-the-art methods in quantifying uncertainty and capturing dynamic correlation for short-term renewable energy scenario generation. The code is open-source and available at: https://github.com/DREAMDXC/Dynamic-Correlation-Quantile-Network

## I. Introduction

The rapid development of renewable energy sources, particularly wind and solar power, introduces significant uncertainty into the power system. Consequently, power system operators must periodically update renewable energy forecasts to support decision-making [1]. Given the inherent chaos of atmospheric systems, probabilistic forecasting is essential for characterizing the uncertainty of renewable energy generation.

For multi-step forecasting problems, a joint probability distribution is required to model the correlations between different variables [2]. Compared to independent marginal probability distributions, the temporal correlations of variables reflect the fluctuations in renewable energy power, which are critical for sequential decision-making processes in power systems, such as unit commitment [3] and economic dispatch [4]. However, many implicit or complex joint probability distribution models do not yield closed-form solutions in system optimization problems. To overcome this challenge, a finite number of temporal scenarios can be sampled to approximate the joint probability distribution in stochastic optimization models [5]. This process, which involves modeling the joint probability distribution and sampling discrete scenarios, is referred to as scenario generation [6].

The atmospheric system, being a nonlinear and time-varying system, exhibits significant variations in correlation across its different phases. Under stable meteorological conditions, renewable energy generation fluctuates less, resulting in stronger temporal correlations. In contrast, during highly fluctuating meteorological conditions, temporal correlations weaken, leading to increased power fluctuations and a higher demand for power system flexibility. Therefore, it is crucial to consider dynamic temporal correlations for renewable energy generation modeling.

Among the methods for modeling joint probability distributions, generative modeling is a rapidly growing branch driven by artificial intelligence technology. The generative models use deep neural networks to map a prior probability distribution to a joint probability distribution. This category mainly includes generative adversarial networks (GANs), variational autoencoders (VAEs), normalizing flows (NFs), and diffusion-based models [7-8]. In the context of renewable energy datasets, generative models are extensively used for data augmentation [9]. When modeled with specific labels as conditional values, generative models can generate corresponding scenarios [10-11] or integrate features from different scenarios [12]. Generative models are highly capable of simulating various data styles. However, when faced with short-term scenario generation task, the crucial goal is to ensure forecast accuracy [13]. This issue naturally leads to the use of proper scoring rules, which compare measured power against generated scenarios [14]. Theoretically, ensuring that a generative network mapping satisfies proper scoring rules requires a reversible network. This has been demonstrated in NF, which rely on restricted network structures, thereby limiting the expressive power of the network [15]. As a result, several studies have combined regression tasks with generative models to enhance performance in short-term scenario generation tasks [13], [16]. While it is undeniable that generative models possess strong capabilities for modeling

This work was supported in part by National Key Research and Development Program of China under Grant 2022YFB2403000.

X. Dong, Y. Lin, X. Zhang *(corresponding author)* and S. Mei are with the State Key Laboratory of Power System Operation and Control, Department of Electrical Engineering, Tsinghua University, Beijing, 100084, China (e-mail: dream_dxc@163.com; lyl23@mails.tsinghua.edu.cn; zhangxuemin@mail.tsinghua.edu.cn; meishengwei@ mail.tsinghua.edu.cn).

implicit temporal dynamic correlations, they face notable challenges regarding training stability and interpretability.

An alternative approach is the use of decoupled models, which model the joint probability distribution by separately estimating the marginal distribution of each variable and the correlation structure between variables. The marginal distribution of a univariate variable serves as the foundation for modeling joint probability distributions [17]. A marginal distribution can be represented by the probability density function (PDF), cumulative distribution function (CDF), or its inverse, the quantile function. Parametric PDF provide continuous uncertainty quantification by making assumptions about the probability distribution for future renewable energy, such as Gaussian distribution, t-location distribution, Weibull distribution and others [18-20]. In practice, there is often a mismatch between the actual and assumed distributions. To improve generalization performance, mixture distribution estimation [21], Bayesian estimation [22], and kernel density estimation [23] are often used for PDF modeling. In addition to PDF, the quantile function is another method for modeling. Quantile regression (QR) can be used to generate non-parametric probabilistic forecasts. Traditional QR models fix a finite number of percentiles, resulting in discrete quantiles [24]. However, discrete quantile regression requires a predefined set of percentiles, making it incapable of modeling quantile functions continuously. During the inverse sampling process for scenario generation, if unknown quantile levels are required, additional spline functions or re-modeling of the quantiles must be used.

Once the marginal probability distributions have been modeled, various approaches can be employed to capture the correlation structure between variables in the joint probability distribution. These methods include time series models, Markov chains, and Copula functions. Time series models use linear regression functions to capture the correlations between variables [25]. Markov chain, on the other hand, uses a non-parametric transfer matrix to model the correlation structure [26]. Higher-order Markov chain and time series models can capture temporal correlations on longer time scales, albeit with increased computational complexity [27]. Copula functions have the ability to connect marginal distribution functions of any form, whether continuous or discrete, parametric or non-parametric [28-29]. Decoupled models, which estimate marginal distributions and correlation structures separately, allow for the application of proper scoring rules, enhancing interpretability and guiding the modeling process effectively. Despite these advancements, most existing methods model correlation structures using static correlation functions, thereby failing to account for the time-varying nature of temporal correlations.

To combine the strengths of generative models and decoupled models, this paper proposes a data-driven dynamic temporal correlation modeling framework for short-term renewable energy scenario generation, with the following key contributions:

**(1) Dynamic Decoupled Mapping Path:** A novel decoupled mapping path is introduced to realize the decoupled transformation from the prior probability distribution to the target joint probability distribution conditioned on covariates. Dynamic correlation network (DCN) models the time-varying characteristics of renewable energy through dynamic covariance matrix.

**(2) Nonparametric Continuous Inverse Sampling:** The proposed implicit quantile network (IQN) implements nonparametric continuous QR, eliminating the dependence on predefined percentiles required in traditional discrete QR models. This innovation resolves the discontinuity issues inherent in applying discrete QR for inverse sampling in renewable energy scenario generation.

The rest of this paper is organized as follows: Section II describes the dynamic temporal correlation modeling framework, Section III describes the structure and algorithms of the dynamic correlation quantile network (DCQN), and Section IV summarizes the case study. Numerical results are given in Section V, and finally, conclusions are given in Section VI.

## II. DYNAMIC TEMPORAL CORRELATION MODELING FRAMEWORK

This section formulates the problem of short-term renewable energy scenario generation and describes the idea of the dynamic temporal correlation modeling framework. The stochastic process of renewable energy generation can be represented by a conditional joint probability distribution $P(\boldsymbol{y} \mid \boldsymbol{x}) = P(y_1, y_2, \ldots, y_T \mid \boldsymbol{x})$, where $\boldsymbol{y}$ is the renewable energy power, $\boldsymbol{x}$ is the numerical weather prediction (NWP) covariate and $T$ is the future target time periods. When the $P(\boldsymbol{y} \mid \boldsymbol{x})$ is sampled once, a temporal scenario $\boldsymbol{s} \sim P(\boldsymbol{y} \mid \boldsymbol{x})$ is obtained, where $\boldsymbol{s} = [s_1, s_2, \ldots, s_T]$. By repeatedly sampling $M$ times, the scenario set $\boldsymbol{S} = \{\boldsymbol{s}^1, \boldsymbol{s}^2, \ldots, \boldsymbol{s}^M\}$ is generated, which approximates the $P(\boldsymbol{y} \mid \boldsymbol{x})$.

According to Sklar's theorem, the joint probability distribution $P(\boldsymbol{y} \mid \boldsymbol{x})$ can be connected using a Copula function $C(\cdot \mid \boldsymbol{x})$ and marginal distributions $P(y_t \mid \boldsymbol{x})$ as:

$$P(y_1, y_2, \ldots, y_T \mid \boldsymbol{x}) = C(P(y_1 \mid \boldsymbol{x}), \ldots, P(y_T \mid \boldsymbol{x}) \mid \boldsymbol{x}) \quad (1)$$

The Copula function captures the correlations between marginal distributions, which dynamically change based on the covariate $\boldsymbol{x}$. Leveraging the decoupling concept from Sklar's theorem, a novel joint probability distribution mapping path is proposed as:

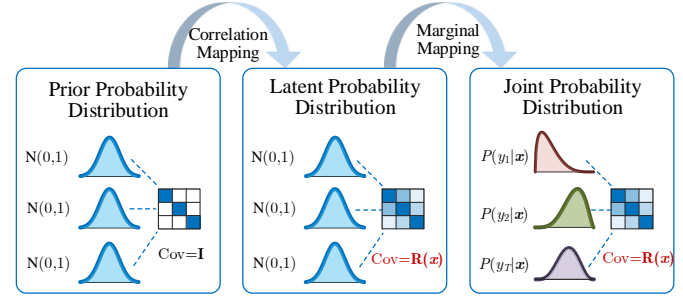

Fig. 1. Mapping paths for the joint probability distribution.

As shown in Fig. 1, the mapping path consists of a two-stage process. Let the prior probability distribution be composed of multiple independent standard Gaussian distributions $z \sim \mathcal{N}(\mathbf{0}, \mathbf{I})$, which have no correlation. The first stage maps the prior probability distribution to the latent probability distribution. In this stage, a correlation structure is introduced to the standard Gaussian distributions without altering the marginal probability distributions. It can be viewed as a correlation transformation into a multivariate Gaussian distribution $z' \sim \mathcal{N}(\mathbf{0}, \mathbf{R}(x))$, where $\mathbf{R}(x)$ is the dynamic covariance matrix. The second stage maps the latent probability distribution to the target joint probability distribution. In this stage, the marginal standard Gaussian distributions $\mathcal{N}(0,1)$ are transformed into the target conditional probability distributions $P(y_t \mid x)$. Through these two stages mapping path, the sample set $Z = \{z^1, z^2, \ldots, z^M\}$ is drawn from the prior distribution is transformed into the corresponding temporal scenario set $S = \{s^1, s^2, \ldots, s^M\}$.

Compared to traditional generative models, the decoupled mapping path offers greater flexibility for scenario generation. It allows for the marginal distribution and correlation structure to be modeled separately by regression model using proper scoring rules. Furthermore, the decoupled mapping explicitly captures the marginal features and correlations between variables, enhancing the interpretability of an otherwise black-box model.

### III. DYNAMIC CORRELATION QUANTILE NETWORK

#### A. DCQN Structure

A DCQN model is developed to implement a dynamic temporal correlation modeling framework using a data-driven approach. For clarity, the structure of the DCQN with two-dimensional scenario mapping process are illustrated in Fig. 2.

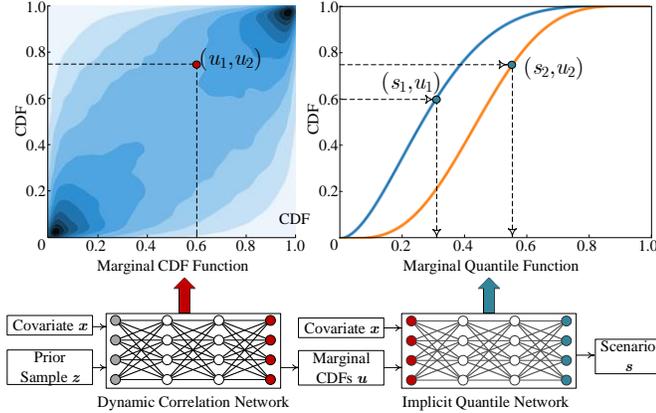

Fig. 2. The structure of DCQN.

The DCQN uses a two-stage mapping process. The first stage involves a dynamic correlation network (DCN), which models the time-varying temporal correlations between variables. The DCN maps the prior distribution sample $z$ and covariate $x$ to the marginal CDF values $u$ as:
$$u = G_c(x, z, \theta_c) \quad (2)$$
where $G_c(\cdot)$ is the mapping function of DCN and $\theta_c$ is the learnable parameters of the DCN.

At this stage, there is no direct mapping of the latent probability distribution samples $z'$, which facilitates the second stage of inverse sampling. However, the latent probability distribution samples $z'$ can be obtained through the inverse CDF function, expressed as:
$$z' = \Phi^{-1}([u_1, u_2, \ldots u_T]) \sim \mathcal{N}(\mathbf{0}, \mathbf{R}(x)) \quad (3)$$
where $\Phi^{-1}(\cdot)$ is the inverse CDF of the standard Gaussian distribution.

Following the correlation mapping, the second stage is the marginal mapping. The implicit quantile network (IQN) maps the marginal CDFs $u$ to the renewable energy scenario $s$ as:
$$s = G_q(x, u, \theta_q) \quad (4)$$
where $G_q(\cdot)$ is the marginal mapping function and $\theta_q$ is the learnable parameters of IQN.

The IQN corresponds to the inverse sampling of the marginal distribution, using the quantile function to transform the latent distribution into the target conditional probability distributions as:
$$s = [F_{y_1}^{-1}(u_1), F_{y_2}^{-1}(u_2), \ldots, F_{y_T}^{-1}(u_T)] \sim P(y \mid x) \quad (5)$$
where $F_{y_t}^{-1}(\cdot)$ is the quantile function.

Thus, once the multivariate marginal CDF values are obtained, scenarios can be sequentially generated via inverse sampling of the marginal distributions. The subsequent subsections provide detailed descriptions of the DCN and IQN structures and their corresponding algorithms.

#### B. Implicit Quantile Network

The traditional discrete QR model provides a quantile value $y^u$ on a CDF value $u$ and is trained by minimizing the quantile loss function, also known as the pinball score (PS):
$$\rho(y, y^u) = \begin{cases} u(y - y^u), & y \geq y^u \\ (u-1)(y - y^u), & y < y^u \end{cases} \quad (6)$$
where $y$ is the measured renewable energy power.

The discrete QR provides fixed CDF values from [0,1]. However, fixing limited CDF values for QR cannot provide a continuous mapping of the quantile function $F^{-1}(\cdot)$. Therefore, the continuous quantile regression is introduced. Fig. 3 visualizes the differences between discrete and continuous quantile regression models.

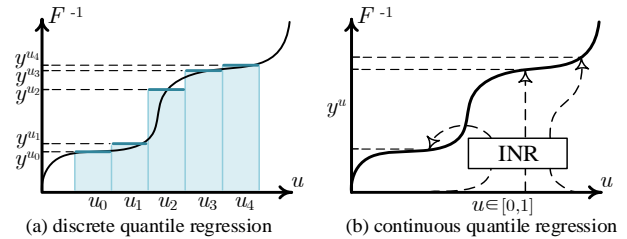

Fig. 3. Comparison of quantile regression models.

In Fig. 3(a), discrete QR cannot describe the tail features of variables and commonly set the $F^{-1}(0)$ and $F^{-1}(1)$ to the natural lower and upper bound. When multiple quantiles need to be estimated, the loss function as:
$$\mathrm{QL}(y, y^{u_i}) = \mathbb{E}_{u_i}[\rho(y - y^{u_i})] \quad (7)$$
where $u_i \in \{u_1, u_2, \ldots, u_I\}$ is a fixed discrete set of CDF values. As the number of CDF values increases, the computational burden of the QR model increases.

In contrast, IQN based continuous quantile regression provides a desired percentile $u$ as an input to neural networks, which is used to fit an implicit neural representation (INR) of the quantile function $y^u = F_\theta^{-1}(u)$, as shown in Fig. 3(b). The loss function is generalized to minimize the quantile divergence as:

$$\text{QD}(y, y^u, u) = \mathbb{E}_{u \sim U[0,1]}[\rho(y - F_\theta^{-1}(u))] \quad (8)$$

The quantile divergence can be trained by a sample drawn from $u \sim U[0,1]$. The sample gradient $\nabla_\theta \rho(y - F_\theta^{-1}(u))$ is an unbiased estimate of quantile divergence, which has been evidenced in reference [30].

Compared to discrete QR models, IQN has the following advantages: IQN does not depend on fixed $u$. By adjusting the novel quantile levels $u$, IQN can output the corresponding $F_\theta^{-1}(u)$ without re-modeling. In addition, continuous QR can be better adapted to inverse sampling $s = F^{-1}(u)$ for scenario generation in the marginal mapping process. The network structure of the IQN is shown in Fig. 4.

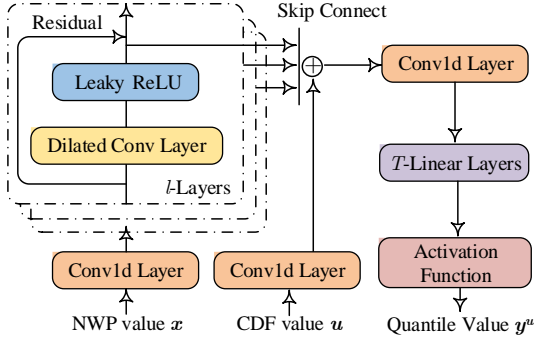

Fig. 4. Structure of IQN.

As can be shown in Fig. 4, the inputs in the IQN are the NWP value $x$ and the CDF value $u \sim U[0,1]$. The output is the quantile value $y^u$. In NWP feature extraction process, the NWP value is passed through $l$-layers temporal convolution modules. Each temporal convolution module consists of a dilated convolution layer and a LeakyReLU activation function. To prevent network degradation, each module is connected by a residual connection. The output values from the various temporal convolution modules are combined using skip connections.

On the output side of the IQN, the channels are downscaled by 1-dimensional convolution and subsequently passed through $T$-linear layers and an activation function. Since renewable energy is normalized and the CDF values fall within the range of [0,1], the Sigmoid function is used in the output layer. It is important to note that the IQN maps the $T$ marginal quantile functions, establishing a one-to-one correspondence between marginal CDF value $u$ and quantile value $y^u$. Therefore, $T$-linear layers are used to map marginal quantile functions.

During the training stage, IQN needs to minimize the expected quantile divergence loss function as:

$$l_{\text{IQN}} = \mathbb{E}[\text{QD}(y, y^u, u)] \quad (9)$$

Once the IQN is trained, a prerequisite for training the DCN is to infer the marginal CDFs $\tilde{u}$ given the measured renewable energy $y$. An effective solution is to use a decoder that is symmetric to the IQN, forming an auto-encoder. This decoder is used to model the inverse mapping process of the IQN [31].

C. Dynamic Correlation Network

The DCN maps the marginal CDF values to describe correlations. Based on the multivariate Gaussian distribution, the prior sample $z$ can be transformed to latent sample $z'$ by a linear transformation formula $z' = Lz$, where $L$ is a lower triangular matrix obtained through the Chebyshev decomposition of the covariance matrix, as $R = LL^\top$. The marginal CDFs $u$ can be obtained by the CDF as $u = \Phi(Lz)$. Therefore, the DCN regresses the lower triangular matrix $L$ to achieve the mapping of marginal CDFs.

The PDF of the multivariate Gaussian distribution as:

$$P(z') = (2\pi)^{-\frac{T}{2}} |R|^{-\frac{1}{2}} \exp[-\frac{1}{2} z'^\top R^{-1} z'] \quad (10)$$

The DCN can be trained by minimizing the negative log-likelihood as:

$$l_{\text{DCN}} = \mathbb{E}_{(\tilde{z}')}[\frac{1}{2}\ln|R| + \frac{1}{2}\tilde{z}'^\top R^{-1}\tilde{z}'] + \text{const} \quad (11)$$

where $\tilde{z}'$ is the multivariate Gaussian distribution value given the marginal CDFs $\tilde{u}$ as $\tilde{z}' = \Phi^{-1}(\tilde{u})$; $|R|$ is the determinant of the covariance matrix, and can be converted by $\ln|R| = 2\ln|L|$.

To regress the lower triangular matrix, the structure of the DCN is shown in Fig. 5.

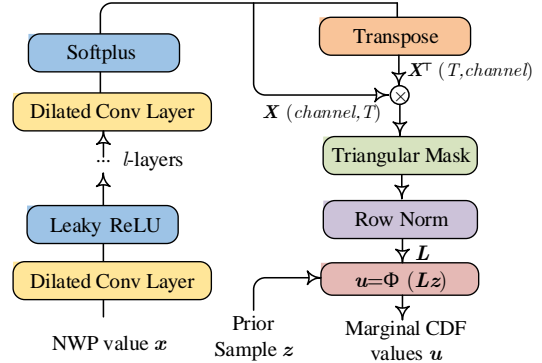

Fig. 5. Structure of DCN.

The NWP value is inputted through $l$-layer temporal convolution modules. The output feature is then passed through the Softplus function to ensure a stable parameterization of the matrix $L$. Furthermore, the matrix $L$ possesses three distinct features, and the DCN is designed for these features in a one-to-one manner.

(1) The matrix $L$ is a $T$-dimensional square matrix, so the channel transpose product layer is designed as:

$$X_{(T,T)} = X_{(T,channel)}^\top X_{(channel,T)} \quad (12)$$

where $X$ is the NWP feature data output by the temporal convolution modules.

(2) The matrix $L$ is a lower triangular matrix, meaning that its upper triangular elements are 0. When the transpose product is performed to obtain a symmetric dense matrix $X$. To ensure consistency, the upper triangular elements of $X$ are converted to 0 using the triangular mask layer.

# REPLACE THIS LINE WITH YOUR PAPER IDENTIFICATION NUMBER (DOUBLE-CLICK HERE TO EDIT) < 5

(3) Since the diagonal of the $R$ matrix is 1, $L$ needs to satisfy the following:

$$R(t,t) = L(t,:)L^\top(t,:) = 1, t \in \{1,2,...,T\} \quad (13)$$

Therefore, a normalization layer is used to normalize each row in $X$ by $L_2$-norm as:

$$L(t,t') = X(t,t') / \sqrt{\| X(t,:) \|^2}, \ t,t' \in \{1,2,...,T\} \quad (14)$$

The above three tricks essentially guarantee that the DCN can regress a valid lower triangular matrix.

### D. Algorithms

During the training process, IQN training can be conducted first, followed by DCN training. This decoupled training is advantageous for applications that only require marginal distributions, as it allows uncertainty modeling to be achieved using only the IQN. In the scenario generation process, the prior distribution samples are sequentially mapped to the scenario using both the DCN and IQN. The pseudo-code for scenario generation is summarized in Algorithm 1.

---

**Algorithm 1:** Scenario generation procedure of the DCQN

**Input:** The NWP data $x$; the number of scenario generation $M$; optimized parameters $\theta_q^*, \theta_c^*$.
**for** scenario number iteration **do**
  prior sample: $z \leftarrow \mathcal{N}(0,1)$
  marginal CDFs generation: $u \leftarrow G_c(x,z,\theta_c^*)$
  scenario generation: $s \leftarrow G_q(x,u,\theta_q^*)$
**end for**
**return** scenario set

---

## IV. CASE STUDY

### A. Data Source Description and Experiment Settings

The dataset was derived from the GEFCom 2014 [32]. Wind and solar data from zones 1 were used for the case study, focusing on day-ahead renewable energy scenario generation. The 24-hour ahead NWP data served as covariates. The renewable energy power data were provided as per-unit values. The dataset spans from January 1, 2012, to December 31, 2013, with a 1-hour time resolution. The 60% dataset was used as the training set, 10% dataset was used as the validation set, while the 30% was used as the test set to evaluate forecasting performance.

The models were implemented on a computer with an NVIDIA TITAN V graphics card and an Intel Core i9-7900X 3.30 GHz. All deep learning models were designed using Pytorch 2.1.0 backend.

### B. Benchmark Model

The case sets up three benchmark decoupled models. The marginal distribution models are multi-horizon quantile recurrent neural network (MQRNN) [33], deep autoregression (DeepAR) [34], and improved deep mixture density network (IDMDN) [21]. MQRNN is a discrete QR model based on a sequence-to-sequence neural networks that uses percentiles from 0.05 to 0.95 with a 0.05 percentile interval. DeepAR uses long short-term memory to fit parametric Gaussian distributions. IDMDN uses a dense network to fit Beta mixture distributions. All mentioned decoupled models use a static Gaussian copula by maximum likelihood estimation for correlation modeling. The QR and PDF models use median and expected values as deterministic forecasts, respectively.

Furthermore, two popular generative models, namely GAN and VAE, were used as benchmark models. The GAN uses the Wasserstein distance as the loss function. The VAE uses a variational lower bound as the loss function. NWP data is used as condition values for these generative models to achieve short-term scenario generation, and the deterministic forecasts are obtained by calculating the expected value of the scenario set.

To ensure equitable comparisons of deep learning models, the number of parameters in the benchmark models is kept at a consistent level to minimize the undue influence of model size differences on the training results. Both the GAN and VAE use the temporal convolution network structure that is similar to the proposed DCQN.

### C. Evaluation Metrics

Multiple evaluation metrics are used to comprehensively evaluate scenario generation task. To differentiate between the evaluation of marginal and joint probability distributions, probabilistic forecasting is treated as the modeling of marginal probability distributions, while scenario generation is regarded as the modeling of joint probability distributions. All metrics are averaged over the test samples.

#### 1) Deterministic Forecasting

The scenario generation task is probabilistic, representing the probability distribution of future events. However, point forecasts remain crucial for practical applications due to the need for decision-making and accuracy assessment.

Mean absolute error (MAE) and root mean square error (RMSE) are used as evaluation metrics for deterministic forecasting as follows:

$$\text{MAE} = \frac{1}{T}\sum_{t=1}^{T}| y_t - \widehat{y}_t | \quad (15)$$

$$\text{RMSE} = \sqrt{\frac{1}{T}\sum_{t=1}^{T}(y_t - \widehat{y}_t)^2} \quad (16)$$

where $\widehat{y}_t$ is the deterministic forecast results; $y_t$ is the measured renewable energy.

#### 2) Probabilistic Forecasting

The PS and continuous ranked probability score (CRPS) are used as evaluation metrics for marginal distribution [35]. The PS is calculated as Eq. (6). The CRPS is calculated as follows:

$$\text{CRPS} = \frac{1}{T}\sum_{t=1}^{T}(\frac{1}{M}\sum_{m=1}^{M}| s_t^m - y_t | - \frac{1}{2M^2}\sum_{m,m'=1}^{M}| s_t^m - s_t^{m'} |) \quad (17)$$

where $s_t^m$ is the $t$-th step renewable energy value of the $m$-th scenario.

#### 3) Scenario Generation

The energy score (ES) and variogram score (VS) are used as evaluation metrics for joint distribution [36]. The VS specifically measures the accuracy of time-varying correlation modeling.

$$\text{ES} = \frac{1}{M} \sum_{m=1}^{M} \| \boldsymbol{s}^m - \boldsymbol{y} \|_2 - \frac{1}{2M^2} \sum_{m,m'=1}^{M} \| \boldsymbol{s}^m - \boldsymbol{s}^{m'} \|_2 \quad (18)$$

where $\boldsymbol{s}^m$ is the $m$-th scenario; $\|\cdot\|_2$ is the Euclidean norm.

$$\text{VS} = \sum_{t,t'=1}^{T} (| y_t - y_{t'} |^{\frac{1}{2}} - \frac{1}{M} \sum_{m=1}^{M} | s_t^m - s_{t'}^m |^{\frac{1}{2}})^2 \quad (19)$$

## V. RESULTS AND DISCUSSION

### A. Deterministic Forecast Results

This section compares the deterministic forecast results of DCQN with benchmark models. Table I presents the MAE and RMSE of the forecast results. The best and second-best values for each evaluation metric are highlighted in bold and underlined, respectively.

TABLE I
DETERMINISTIC FORECAST RESULTS

| Models | Wind | | Solar | |
| --- | --- | --- | --- | --- |
| | MAE | RMSE | MAE | RMSE |
| MQRNN | 0.1220 | 0.1524 | 0.0331 | 0.0645 |
| DeepAR | 0.1266 | 0.1544 | 0.0397 | 0.0744 |
| IDMDN | 0.1221 | 0.1499 | 0.0353 | 0.0654 |
| VAE | 0.1372 | 0.1642 | 0.0479 | 0.0752 |
| GAN | 0.1319 | 0.1622 | 0.0443 | 0.0778 |
| DCQN | **0.1176** | **0.1474** | **0.0320** | **0.0616** |

The deterministic forecast results indicate that the DCQN model outperforms other models in both wind and solar forecasting. Among the decoupled models, MQRNN and IDMDN follows closely, with slightly higher errors than DCQN. The non-parametric estimation of MQRNN exhibits lower metrics, as the quantile loss function is equivalent to the MAE loss function at $u = 0.5$. In contrast, VAE and GAN models exhibit the highest errors across both wind and solar forecasitng, highlighting their relative limitations in deterministic forecast regression task.

### B. Probabilistic Forecast Results

This section compares the probabilistic forecast results of DCQN with benchmark models. Table II lists the PS and CRPS results, while Fig. 6 and Fig. 7 shows the day-ahead probabilistic forecast results of the three consecutive days in the test set. The figure plots the probabilistic forecasts using 10% to 90% confidence intervals.

TABLE II
PROBABILISTIC FORECAST RESULTS

| Models | Wind | | PV | |
| --- | --- | --- | --- | --- |
| | PS | CRPS | PS | CRPS |
| MQRNN | 0.0451 | 0.0862 | 0.0128 | 0.0248 |
| DeepAR | 0.0473 | 0.0904 | 0.0148 | 0.0284 |
| IDMDN | 0.0447 | 0.0853 | 0.0130 | 0.0247 |
| VAE | 0.0549 | 0.1057 | 0.0210 | 0.0412 |
| GAN | 0.0493 | 0.0944 | 0.0180 | 0.0350 |
| DCQN | **0.0444** | **0.0849** | **0.0121** | **0.0232** |

The results indicate that DCQN achieves the best performance across both metrics, followed by MQRNN and IDMDN. Among the decoupled models, the parametric method DeepAR demonstrates lower accuracy compared to non-parametric models, owing to the greater flexibility offered by non-parametric approaches. Unlike deterministic forecasting, IDMDN outperforms MQRNN, which can be attributed to its ability to model distributions continuously. On the other hand, the performance metrics of generative models are inferior to those of decoupled models, particularly for VAE. As shown in Fig. 6(d) and Fig. 7(d), the confidence interval of VAE is narrower and less reliable due to the approximate computation of the reconstruction loss term in the evidence lower bound. In contrast, DCQN leverages a quantile divergence loss function for training marginal distributions, offering significant advantages over VAE and GAN by explicitly incorporating probability constraints into the modeling of marginal distributions.

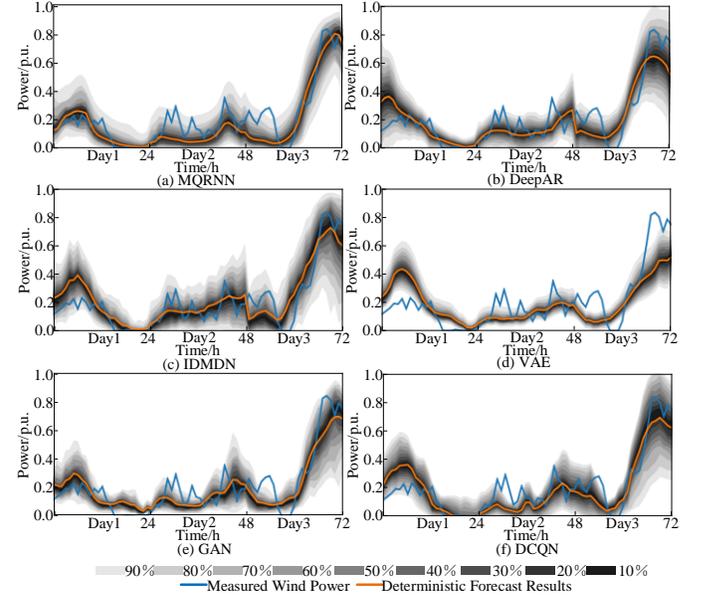

Fig. 6. Day-ahead wind power probabilistic forecast results.

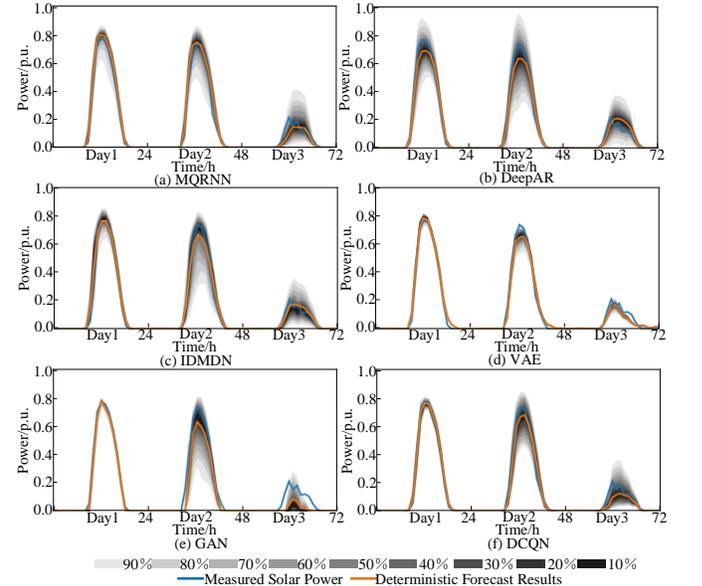

Fig. 7. Day-ahead solar power probabilistic forecast results.

### C. Scenario Generation Results

This section analyzes the results of scenario generation, which requires consideration of the temporal correlations between the variables. Table III shows the ES and VS results. To visualize the scenario generation results, Fig. 8 and Fig. 9 shows the day-ahead scenarios of the three consecutive days in the test set.

TABLE III
SCENARIO GENERATION RESULTS

| Models | Wind | | Solar | |
|---|---|---|---|---|
| | ES | VS | ES | VS |
| MQRNN | 0.5308 | 16.399 | 0.2494 | 5.628 |
| DeepAR | 0.5411 | 17.370 | 0.2619 | 5.840 |
| IDMDN | 0.5269 | 16.133 | 0.2314 | 4.850 |
| VAE | 0.6251 | 19.298 | 0.3141 | 7.098 |
| GAN | 0.5779 | 17.182 | 0.2883 | 9.324 |
| DCQN | **0.5176** | **16.096** | **0.2177** | **4.381** |

As shown in Table III, DCQN achieves optimal performance across several metrics, with IDMDN closely following. The lower VS values indicate that DCQN effectively captures time-varying correlations. In contrast, VAE and GAN underperform compared to decoupled models in terms of the evaluated metrics. This discrepancy can be attributed to the uncertainty modeling bias inherent in traditional generative models, which fail to satisfy proper scoring rules. Consequently, incorporating regression tasks is essential for improving the performance of generative models in short-term scenario generation.

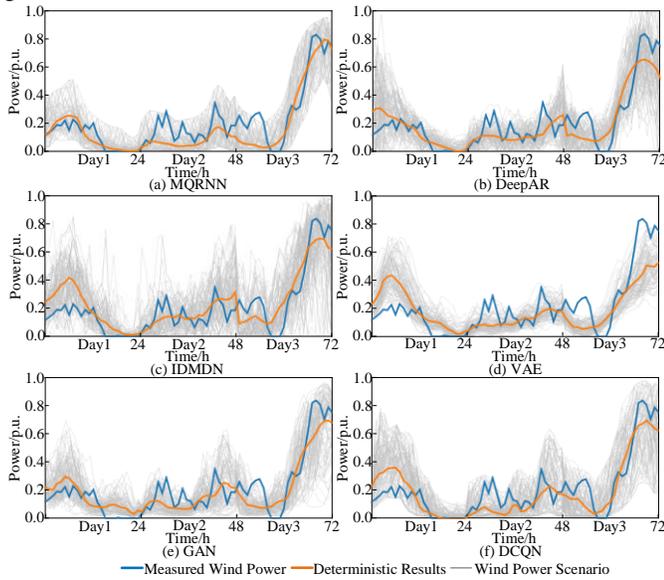

Fig. 8. Day-ahead wind power scenario generation results.

In Fig. 8 and 9, the results of MQRNN were post-processed to be restricted between $\widehat{y}^{0.05}$ and $\widehat{y}^{0.95}$ which aims to prevent unusual tail sampling values. This adjustment is necessary due to the discrete QR employed by MQRNN. While MQRNN regresses quantiles ranging from 0.05 to 0.95, it fails to adequately model the tail features of the marginal distribution. During scenario generation, the values within the CDF intervals [0,0.05] and [0.95,1] are uniformly distributed between $[0,\widehat{y}^{0.05}]$ and $[\widehat{y}^{0.95},1]$ through inverse sampling. This leads to abnormal values in the generated scenarios. Thus, compared to discrete QR, IQN is better able to model the tail features of the distribution through continuous QR. A similar issue arises in DeepAR. In parametric modeling of marginal distributions, the Gaussian distribution used by DeepAR is unbounded, which can lead to scenario values exceeding the [0,1] range, resulting in density leakage. Consequently, the generated scenarios require clipping to remain within the effective boundaries. In contrast, IDMDN and IQN are inherently bounded within the [0,1] range, making them better suited for fitting marginal probability distributions to guide renewable energy scenario generation. However, due to the tail behavior of the beta distribution, IDMDN exhibits slight nighttime power when generating solar scenarios.

In terms of volatility in renewable energy scenarios, decoupled models are more prone to exhibiting spikes in temporal scenarios due to their reliance on static correlation modeling. This tendency is more pronounced in DeepAR and IDMDN, while MQRNN mitigates this issue through preprocessing. On the other hand, generative models, such as VAEs and GANs, dynamically learn temporal correlations based on covariates, resulting in temporal scenario fluctuations that align more closely with measured renewable energy power. This characteristic makes generative models superior to decoupled models in capturing temporal correlation. DCQN integrates the strengths of both approaches, offering robust performance by balancing dynamic correlation modeling with accurate marginal distribution fitting.

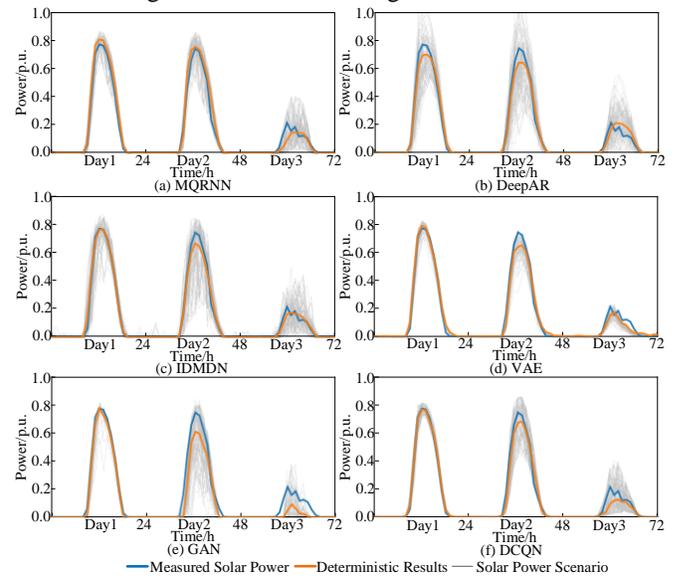

Fig. 9. Day-ahead solar power scenario generation results.

### D. Discussion on Temporal Correlation

This section examines the modeling of time-varying temporal correlation. In order to maintain the marginal distribution, the comparison method uses the static copula function through maximum likelihood estimation as a replacement for the DCN. The covariance matrix is utilized to analyze the temporal correlation. Consistent with the previous case study, the three test days are used for visualization analysis, as shown in Fig. 10 and Fig. 11.

Fig. 10(a) shows that the static covariance matrix reveals high correlations among neighboring variables. However, the correlation significantly decreases when the time intervals between neighboring moments exceed 3 hours. This finding aligns with prior knowledge of wind power, which exhibits a stronger correlation among neighboring moments. Since the static correlation employs the same correlation matrix for all test days, there is no time-varying correlation.

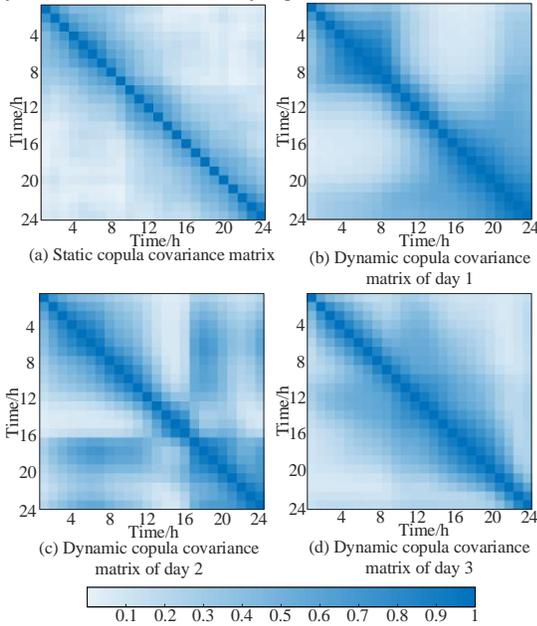

Fig. 10. Heatmap of the wind covariance matrix.

Combined with Fig. 8(f), a comprehensive analysis of the covariance matrix is presented. It is evident that the wind power forecast results on the first day can be divided into three stages: the first stage (1:00 - 8:00) exhibits an upward climbing trend, the second stage (9:00 - 18:00) shows a downward climbing trend, and the third stage (19:00 - 24:00) indicates a gradual decrease in wind power towards zero. The dynamic covariance matrix, shown in Fig. 10(b), demonstrates a stronger correlation during the first stage. Subsequently, the correlation significantly decreases during the rapid downward climb in the second stage. Finally, during the third stage, when wind power stabilizes, the correlations exhibit a higher value.

The same analysis can be applied to the other test days. Combining Fig. 10(c) with Fig. 8(f), on the second day, the correlation decreases when fluctuations occur in the forecast results from 12:00 - 16:00. Similarly, combining Fig. 10(d) with Fig. 8(f), the forecast results on the third day display a downward climb followed by a sustained upward climb, resulting in higher correlations among neighboring moments.

Solar power generation exhibits a clear daily periodicity. As shown in Fig. 11, the temporal correlation of solar power generation shows high similarity. During daytime hours, the power generation on the first day remains relatively stable, with a high correlation between adjacent hours. In contrast, on the second and third days fluctuates significantly, particularly on the third day. This increased variability leads to a marked reduction in the correlation between consecutive time periods. The static correlation matrix fails to capture dynamic features.

In general, when renewable energy power forecast results are stable, the correlation of variables increases during this stage. Conversely, when the renewable energy power forecast results fluctuate or climb, the correlation of variables significantly decreases during the stage. This analysis highlights that the static correlation solely reflects the inherent correlation between variables and does not account for time-varying correlations. While traditional generative models can model time-varying correlations, they are unable to reflect the correlation between variables through visualization of the results. In contrast, the DCN can flexibly model time-varying correlations and provide enhanced interpretability for neural networks.

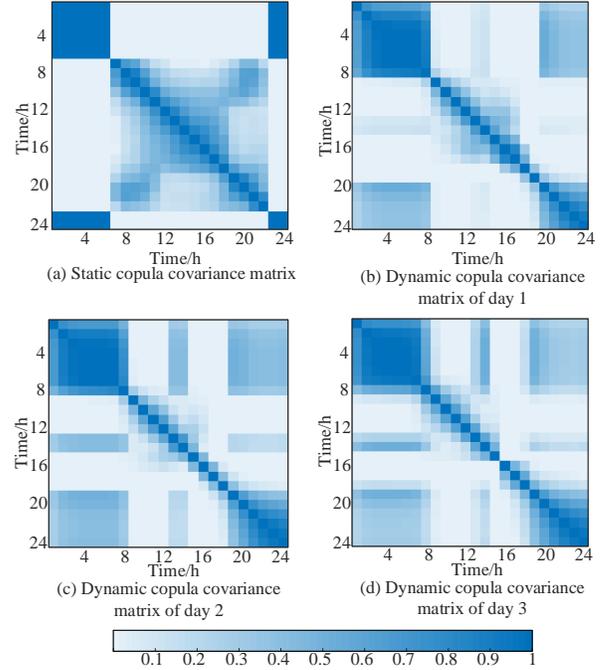

Fig. 11. Heatmap of the solar covariance matrix.

## VI. CONCLUSION

This study proposes a data-driven dynamic temporal correlation modeling framework for short-term renewable energy scenario generation. The scenario generation process is decoupled into two stages: correlation mapping and marginal mapping.

We designed DCQN model combines a DCN and an IQN based on the proposed framework. It offers several advantages over existing models. Unlike discrete QR models, it enables continuous, non-parametric QR and effectively models the tails of marginal distributions. This continuous QR approach is particularly suited for marginal inverse sampling in scenario generation. Furthermore, the DCQN is trained using proper scoring rules, enhancing the accuracy of short-term scenario generation. Additionally, DCN models dynamic temporal correlation between variables, with interpretability further improved by representing the covariance matrix explicitly. A case study utilizing data from the GEFCom 2014 demonstrated the superiority of the proposed model.

While the proposed framework successfully models time-varying correlations, it is constrained by the assumptions of the multivariate Gaussian distribution. Future research should focus on exploring generalized non-parametric correlation modeling approaches using deep neural networks.

**Xiaochong Dong** received the B.S. degree from China Three Gorges University, Yichang, China, in 2017, the M.S. degree in North China Electric Power University, Beijing, China, in 2020, and the Ph.D. degree in North China Electric Power University, Beijing, China, in 2024, both in electrical engineering. He is currently a postdoctoral researcher at Tsinghua University, Beijing, China. His research interests include artificial intelligence and optimization of the power system.

**Yilin Liu** received the B.S. degree in electrical engineering from Tsinghua University, Beijing, China, in 2023. She is currently pursuing a M.S. degree in electrical engineering at Tsinghua University. Her research interests include renewable energy forecast and artificial intelligence.

**Xuemin Zhang** received the B.Sc. and Ph.D. degrees in electrical engineering from Tsinghua University, Beijing, China, in 2001 and 2006, respectively. She is currently an Associate Professor with Tsinghua University. Her research interests include power system stability and control, renewable energy forecast, and power system complexity.

**Shengwei Mei** received the B.Sc. degree in mathematics from Xinjiang University, Urumqi, China, in 1984, the M.Sc. degree in operations research from Tsinghua University, Beijing, China, in 1989, and the Ph.D. degree in automatic control from the Chinese Academy of Sciences, Beijing, in 1996. He is currently a Professor with Tsinghua University. His research interests include power system analysis and control, robust control of power systems, comprehensive utilization of new energy, and disaster prevention of large power grids.